\documentclass[10pt,journal,compsoc]{IEEEtran}

\ifCLASSOPTIONcompsoc
  \usepackage[nocompress]{cite}
\else
  \usepackage{cite}
\fi

\ifCLASSINFOpdf

\else

\fi

\usepackage{graphicx}
\usepackage{epstopdf}
\usepackage{booktabs}
\usepackage{subfigure}
\usepackage{amsfonts}
\usepackage{amsmath}
\usepackage{algorithm, algorithmic}
\usepackage{subfigure}
\usepackage{cite}
\usepackage{amsthm}
\usepackage[T1]{fontenc}
\usepackage[utf8]{inputenc}
\usepackage{authblk}
\usepackage{float}
\usepackage{hyperref}
\usepackage{multirow}
\usepackage{makecell}
\hypersetup{hypertex=true,
            colorlinks=true,
            linkcolor=blue,
            anchorcolor=blue,
            citecolor=blue}

\begin{document}
\title{Metric Distribution to Vector: Constructing Data Representation via Broad-Scale Discrepancies}

\author[1,2]{Xue Liu}
\author[3,4]{Dan Sun}
\author[2]{Xiaobo Cao}
\author[2]{Hao Ye}
\author[3,4,1*]{Wei Wei \thanks{Corresponding author: weiw@buaa.edu.cn}}
\affil[1]{Institute of Artificial Intelligence, Beihang University, Beijing, 100191, P.R.China}
\affil[2]{Beijing System Design Institute of Electro-Mechanic Engineering, Beijing, 100854, P.R.China}
\affil[3]{School of Mathematical Sciences, Beihang University, Beijing, 100191, P.R.China}
\affil[4]{Key Laboratory of Mathematics, Informatics and Behavioral Semantics, Ministry of Education, 100191, P.R.China}

%\author{
%	\IEEEauthorblockN{
%		Xue Liu\IEEEauthorrefmark{1,2},
%		Dan Sun\IEEEauthorrefmark{3,4},
%		Xiaobo Cao\IEEEauthorrefmark{2},
%		Hao Ye\IEEEauthorrefmark{2}
%		and Wei Wei\IEEEauthorrefmark{3,4,1}\thanks{Corresponding author: weiw@buaa.edu.cn}}\\
%
%	\IEEEauthorblockA{\IEEEauthorrefmark{1}Institute of Artificial Intelligence, Beihang University, Beijing, 100191, P.R.China}\\
%	\IEEEauthorblockA{\IEEEauthorrefmark{2}Beijing System Design Institute of Electro-Mechanic Engineering, Beijing, 100854, P.R.China}\\
%	\IEEEauthorblockA{\IEEEauthorrefmark{3}School of Mathematical Sciences, Beihang University, Beijing, 100191, P.R.China}\\
%	\IEEEauthorblockA{\IEEEauthorrefmark{4}Key Laboratory of Mathematics, Informatics and Behavioral Semantics, Ministry of Education, 100191, P.R.China}
%}

\markboth{Preprint submitted to Journal of IEEE TRANSACTIONS ON XXX, 2022}%
{Shell \MakeLowercase{\textit{et al.}}: Bare Demo of IEEEtran.cls for Computer Society Journals}

\IEEEtitleabstractindextext{%
\begin{abstract}
Graph embedding provides a feasible methodology to conduct pattern classification for graph-structured data by mapping each data into the vectorial space. Various pioneering works are essentially coding method that concentrates on a vectorial representation about the inner properties of a graph in terms of the topological constitution, node attributions, link relations, etc. However, the classification for each targeted data is a qualitative issue based on understanding the overall discrepancies within the dataset scale. From the statistical point of view, these discrepancies manifest a metric distribution over the dataset scale if the distance metric is adopted to measure the pairwise similarity or dissimilarity. Therefore, we present a novel embedding strategy named $\mathbf{MetricDistribution2vec}$ to extract such distribution characteristics into the vectorial representation for each data. We demonstrate the application and effectiveness of our representation method in the supervised prediction tasks on extensive real-world structural graph datasets. The results have gained some unexpected increases compared with a surge of baselines on all the datasets, even if we take the lightweight models as classifiers. Moreover, the proposed methods also conducted experiments in Few-Shot classification scenarios, and the results still show attractive discrimination in rare training samples based inference.
\end{abstract}

}

\maketitle

\IEEEraisesectionheading{\section{Introduction}\label{sec:introduction}}
%\renewcommand{\thefootnote}{}
%\footnotetext{The source code and datasets: XXXXXXXXXXXXXXXXXXXX.}
%\footnotetext{The preprint manuscript: XXXXXXXXXXXXXXXXXXXX.}

\IEEEPARstart{G}{raph}, as a kind of structural data, is widely used to represent interactions among units in a relational system, such as a molecule, a social network, or a biological group. In a graph, edges are associated with the connectivity of nodes, defining a topological relationship such as proximity, adjacency, etc. Usually, every graph is assigned a label according to its underlying properties or applications. Thus, inferring the category of a structural graph within an extensive collection of graphs constitutes a significant task in the research of pattern analysis.

\begin{figure*}[h]
\centering
\includegraphics[height=6cm,width=16cm]{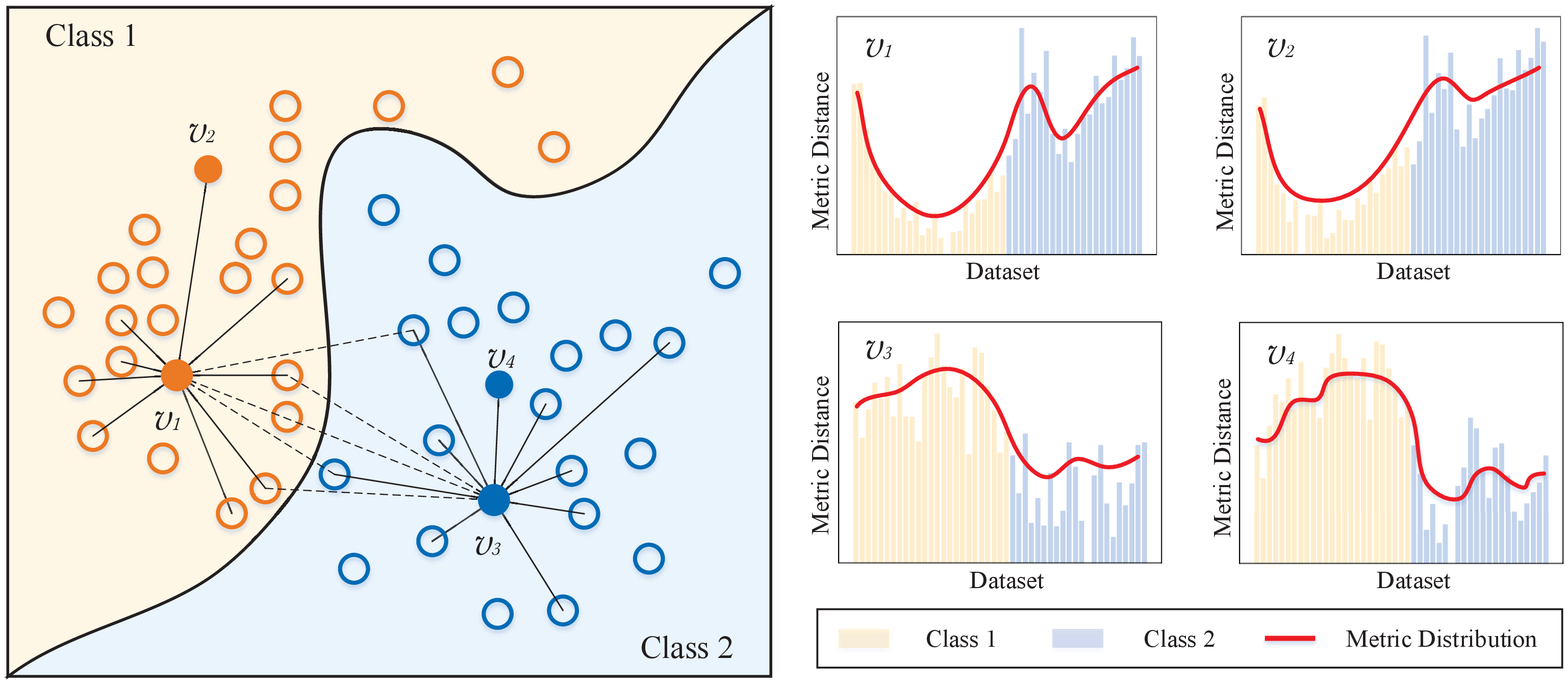}
\caption{A classification example in two dimensions to illustrate the metric distribution. In the scatter plot, each point is labeled with an allocated class denoted by binary colors (i.e., blue and orange). All the instances are separated by a segmentation boundary, denoted as the dark curve. In each class, we use dark lines to denote the distance between intra-group points and use dark dotted lines to represent the distance between inter-group points. In particular, we take the Euclidean distance as the metric in this case. In addition, the metric distributions for $v_{1}$, $v_{2}$, $v_{3}$, and $v_{4}$ are also shown. Among these four data instances, $v_{1}$ and $v_{2}$ are in one class, while $v_{3}$ and $v_{4}$ belong to another category. In each subplot, the histogram reports the distance between the targeted point with each instance (colored according to its class) within the dataset. The red curve exhibits the overall metric distribution trend. The data belonging to the same class clearly possess approximate metric distance distributions.
}
\label{Figure-Metric-Distribution}
\end{figure*}

To handle the pattern classification task, graph embedding methods have provided a feasible way to feed structural data into models like classifiers, predictors, etc. The taxonomy of pioneering graph embedding techniques proposed in the literature could be roughly split into three broad categories: Factorization, Substructure mining, and Deep Learning methods. The Factorization methods, such as GraRep~\cite{GraRep}, Laplacian Eigenmaps~\cite{Laplacian-eigenmaps}, Graph Factorization~\cite{Graph-factorization}, et al., tend to obtain the embedding that more reflects intrinsic geometric features of a graph by factorizing its connection matrix (including adjacent matrix, Laplacian matrix, Katz similarity matrix, etc.) based on different matrix properties. The Substructure methods embed each graph using sampled substructures on graphs to approximate the distribution of nodes, edges, or motifs, and DeepWalk~\cite{Deepwalk}, node2vec~\cite{node2vec}, and AWE~\cite{AWE} et al. are famous techniques. The Deep Learning methods iteratively aggregate the embeddings of each neighbor for a node and then obtain the new embedding of each node as well as the embedding for each graph~\cite{GCN}.

Graph embedding is exactly a task-driven encoded method in general. That is, the adopted graph embedding method should be tightly associated with the targeted machine learning task. In particular, the core issue for pattern classification is to discriminate each object from others and sort each object into different categories according to its global relations (including similarities and dissimilarities) inside the database range. However, various graph embedding techniques are essentially a representation method, typically concentrating on a vectorial reflection about the inner properties of a graph concerning the topological constitution, node attributions, link relations, etc., failing to unravel the global interactions in similarity within the entire dataset. Thus, how to embed a structural graph with fully exhibiting global similarities or dissimilarities deserves much attention.

Calculating the pairwise distance metric is the most fundamental method to measure the discrepancy between the targeted graph and others. Recently, the distance metric has become an indispensable technique to measure pairwise similarity in many successful machine learning models, including k-Nearest Neighbors (kNN)~\cite{KNN}, k-means~\cite{K-Means}, Support Vector Machines (SVMs)~\cite{SVM}, Shortest-Path Kernel~\cite{Shortest-Path-kernel}, Hash Graph Kernel~\cite{Hash-Graph-kernel}, Graph Edit Distance (GED)~\cite{GED-PR, GED-TKDE}, etc. Among these metric-based learning approaches, many attempt to keep all pairwise intra-group points close and separate all pairwise inter-group points far apart.

Noting that the notion of distribution has well described the proportion and dispersion of elements in many fields of science ranging, if we broaden the concentration from pairwise similarities to dataset-scale discrimination, a series of distance metrics associated with the targeted graph actually capture a metric distribution to reveal more global semantics for each entity. In general, this characteristic could lead to a notable improvement in discrimination and interpretation. We take Figure~\ref{Figure-Metric-Distribution} as a concrete instance to illustrate the basic idea. In Figure~\ref{Figure-Metric-Distribution}, sample pairs $(v_{1}, v_{2})$ and $(v_{3}, v_{4})$ are separated into different classes because pairwise nodes in same class possess not only the smaller Euclidean distance but also a similar metric distribution.

Based on above discussion, we propose a graph embedding framework Metric Distribution to vector (\textbf{MetricDistribution2vec}) for the pattern classification task. We emphasize MetricDistribution2vec due to its wide-ranging applicability and generalization for both supervised and Few-Shot classification scenes. Concretely, we target the metric distribution as an embedded discrimination characteristic for each data within the entire database. To measure the similarity between structural graphs, a new pairwise graph distance is presented following the optimal transportation theory. Each graph is fragmented into a mesoscopic decomposition by a set of selected higher-order subgraphs to simulate the generation distribution that pends transporting. Accordingly, the minimum transportation cost would be regarded as the distance between pairwise graphs. We conduct 12 real-world supervised classification problems to validate the effectiveness with 10-fold stratified cross-validation and Few-Shot settings. The experimental results have shown that our method could successfully strengthen advances in accurate prediction on all datasets. On 10-fold stratified cross-validation classification problems, our approach has achieved at least $90\%$ accuracy on all 12 datasets, at least $97\%$ accuracy on 11 datasets, and at least $99\%$ accuracy on 7 datasets, respectively. When $3\% - 8\%$ data are used for training on Few-Shot classification problems, our method exceeds the best baseline on 10 datasets. Once the training data sampling rate rises to $20\%$, our method can outperform all the baselines on all 12 datasets. And in particular, it achieves more than $90\%$ accuracy on 10 datasets.

Specifically, the main contributions are highlighted as follows.
\begin{itemize}
\item[(1)] We propose a novel and scalable framework of graph representation by embedding metric distribution as a globally discriminative characteristic for each structural data in the view of dataset scale.
\item[(2)] We work on the optimal transportation distance to measure the discrepancy between pairwise graphs, which are represented as subgraph decompositions to gain on the ground truth generation distributions.
\item[(3)] Extensive experiments are conducted on several real-world datasets to show the effectiveness in supervised classification as well as Few-Shot scenarios. On most datasets, even only $3\%-8\%$ labeled data are pretty enough to lead a highly accurate, efficient, and lightweight classifier to derive a superior performance on the test set than all other baselines.
\end{itemize}

The rest part of this paper is sketched in the following manner. The basic problems of supervised and Few-Shot classification are stated in Section~\ref{Problem-Statement}. In this paper, we focus on a distribution-based representation for structural data. Correspondingly, the main idea and framework of MetricDistribution2vec are briefly introduced in Section~\ref{MetricDistribution2vec}. As for the graph distance, we present a particular distance metric derived from optimal transportation theory in Section~\ref{optimal-transportation}. The basic concepts including optimal transportation, Wasserstein distance, and the optimal transportation between structural graphs are all covered in this Section. The results of the comparison experiments on both supervised and Few-Shot classification are provided in Section~\ref{Experiments}. Beyond that, the reason for the proposed method suitable for Few-Shot classification is also discussed in Section~\ref{Experiments}. In the last, the main conclusions are summarized in Section~\ref{Conclusion-and-Future-Outlook}.

\section{Problem Statement}\label{Problem-Statement}
In this paper, we concentrate on the general graph embedding problem and its applications in both supervised and Few-shot classification.

Let $\mathcal{G}$ denote an $N$ graph set
\begin{equation}\label{graph-dataset}
  \mathcal{G} = \{G_{i}, G_{2}, \ldots, G_{N}\},
\end{equation}
with each graph as $G_{i} = \{V_{i}, E_{i}\}$, where $V_{i}$ and $E_{i}$ are the node set and edge set. Let $\mathcal{Y} = \{y_{1}, \ldots, y_{K}\}$ be the label space, containing all the $K$ classes of labels for $\mathcal{G}$. Each graph $G_{i} \in \mathcal{G}$ is allocated with a label in $\mathcal{Y}$.

For the input of a graph $G \in \mathcal{G}$ and a preset dimensionality of parameter $d$, the problem of \textbf{graph embedding} is to transform $G$ into a $d-$dimensional vector, in which specific properties (such as topological constitution, node attributions, link relations, etc.) are preserved as much as possible, such that the structural data is easier to be recognized with classifiers to conduct classification.

If we randomly select $S$ instances by sampling ratio $\eta$ from $\mathcal{G}$ into the training set $\mathcal{G}^{Train}$ and randomly select $T$ instances by sampling ratio $\zeta$ into the test set $\mathcal{G}^{Test}$, satisfying
\begin{equation}\label{seperation-training-test}
    \mathcal{G}^{Train} \cap \mathcal{G}^{Test} = \emptyset,\ \mathcal{G}^{Train} \cup \mathcal{G}^{Test} \subseteq \mathcal{G},\ and \ \eta + \zeta \leq 1.
\end{equation}

The issue of \textbf{supervised classification} is to learn a function $\mathcal{F}: \mathcal{G} \rightarrow \mathcal{Y}$ that predicts the label for new input based on the labeled data in the training set. The input to $\mathcal{F}$ will be any instance $G \in \mathcal{G}$, and the output will be the label prediction $\mathcal{F}(G) \in \mathcal{Y}$.

Supervised classification has been highly successful in data-intensive cases but is often hampered when sufficient examples are hard or impossible to acquire. To tackle this problem, \textbf{Few-Shot classification}~\cite{few-shot-classification} has attracted much attention in recent years. Few-Shot Learning aims to augment the supervised experience with the prior knowledge learned from only a few samples to new tasks. Concretely, Few-Shot Classification learns classifiers given only on a few labeled examples of each class in training set $\mathcal{G}^{Train}$. In this paper, we consider the general $K$-Way Few Shot classification~\cite{K-way-few-shot}, in which $\mathcal{G}^{Train}$ sampled with a small sampling rate $\eta$ containing $S$ instances from $K$ classes.

\section{Representation by global metric distribution}\label{MetricDistribution2vec}
The graph embedding techniques have provided a feasible way to apply structured graph data into machine learning models. Here, we illustrate a graph embedding framework by metric distribution as a discriminative property for each graph.

\subsection{Metric Distribution}

For $K$-class classification, we consider the similarity partition of graph dataset $\mathcal{G}$ as a union of $K$ disjoint equivalence sets,
\begin{equation}\label{G}
  \mathcal{G} = \bigcup\limits_{k = 1}^{K} \mathcal{S}^{(k)},
\end{equation}
in which $\mathcal{S}^{(k)}$, $k=1, \ldots, K$, is denoted as a set of graphs allocated with label $y_{k}$:
\begin{equation}\label{equivalence-set}
  \mathcal{S}^{(k)} = \{ G_{1}^{(k)}, \ldots, G_{N_{k}}^{(k)} \}.
\end{equation}

We rearrange graphs in $\mathcal{G}$ by looking at the order of labels in set $\mathcal{Y}$ from $1$ to $K$ as
\begin{equation}\label{sampled-G-rearrange}
  \mathcal{G} = \{G_{1}^{(1)}, \ldots, G_{N_{1}}^{(1)}, \ldots , G_{1}^{(K)}, \ldots, G_{N_{K}}^{(K)} \}.
\end{equation}

If we take a real-valued function $d$ on graph Cartesian product $\mathcal{G} \times \mathcal{G}$ as the metric function,
\begin{equation}\label{metric-distance-function}
  d: \mathcal{G} \times \mathcal{G} \rightarrow [0, +\infty),
\end{equation}
for graph $G_{i} \in \mathcal{G}$, the metrics corresponding to equivalence set $\mathcal{S}^{(k)}$ are orderly stacked in a $1 \times N_{k}$ vector
\begin{equation}\label{metric-distribution-single-equivalence-set}
  M_{G_{i}}^{(k)} = [d_{i,1}^{(k)},\ldots, d_{i,N_{k}}^{(k)}],
\end{equation}
where each element $d_{i,j}^{(k)} = d(G_{i}, G_{j}^{(k)})$, $j=1,\ldots, {N}_{k}$ denotes the distance between $G_{i}$ and $G_{j}^{(k)}$.

The collection of metrics for $G_{i}$ with respect to ordered $\mathcal{S}^{(k)}$, $k = 1, \ldots, K$, is concatenated as a $1 \times N$ vector
\begin{equation}\label{metric-distribution}
\begin{aligned}
    M_{G_{i}}   & = concatenated(M_{G_{i}}^{(1)}, \ldots, M_{G_{i}}^{(K)}) \\
                & = [d_{i,1}^{(1)}, \ldots, d_{i,N_{1}}^{(1)}, \ldots, \ d_{i,1}^{(K)}, \ldots, d_{i,N_{K}}^{(K)}]
\end{aligned}
\end{equation}

Then, we can define a normalized scheme to represent the \textbf{metric distribution} for $G_{i}$ by a $1 \times N$ vector
\begin{equation}\label{norm-metric-distribution}
\begin{aligned}
  M_{G_{i}} = [\frac{d_{i,1}^{(1)}}{\sigma_{i}}, \ldots, \frac{d_{i,N_{1}}^{(1)}}{\sigma_{i}}, \ldots, \frac{d_{i,1}^{(K)}}{\sigma_{i}}, \ldots, \frac{d_{i,N_{K}}^{(K)}}{\sigma_{i}}]_{1 \times N}.
\end{aligned}
\end{equation}
where $\sigma_{i} = \sum_{k = 1}^{K} \sum_{j = 1}^{N_{k}} d_{i,j}^{(k)}$.

\subsection{Metric distribution to vector: a graph embedding mode for supervised classification}
With the definition described above, we would like to embed the metric distribution of each graph into its vectorial representation. We present the method of \textbf{metric distribution to vector} (MetricDistribution2vec) and its application in supervised classification.

In formula~(\ref{norm-metric-distribution}), there are a number of limitations that one has to attend to in practice:
\begin{itemize}
\item Formula~(\ref{norm-metric-distribution}) uses all the distances associated with both training and test objects to simulate the metric distribution for each sample. However, in real-world supervised scenarios, the test samples are always unknown until they are fed to the predictor for the test, so that it is not feasible to globally understand all pairwise distances from the entire database in advance.
\item It is also expensive of computing to derive all the metric distribution for each graph with computational complexity $\emph{O}[N^{2} \cdot \emph{O}(d)]$, due to at least $\frac{1}{2}(N^2 + N)$ metric distributions pending calculation, where $N$ denotes the graph number and $\emph{O}(d)$ is the computational complexity for metric function $d$.
\end{itemize}

As there seems unnecessary to approximate the metric distribution for every graph using the entire dataset, the typical way is to put down a reasonable set of training samples as the support set and then represent each one in both the training and test set relying only on its relations with the training set. We generate the representation method with the following steps.

Concretely, we first randomly select $\tilde{N}$ training items from $\mathcal{G}$ by sampling rate $\eta$ to form the support set $\mathcal{G}^{Train}$, which could be represented as $K$ disjoint segmentations,
\begin{equation}\label{G-training}
\begin{aligned}
  \mathcal{G}^{Train}   & = \bigcup\limits_{k = 1}^{K} \mathcal{\tilde{S}}^{(k)} \\
                                & = \{G_{1}^{(1)}, \ldots, G_{\tilde{N}_{1}}^{(1)}, \ldots , G_{1}^{(K)}, \ldots, G_{\tilde{N}_{K}}^{(K)} \},
\end{aligned}
\end{equation}
in which $\tilde{N} = \sum_{k=1}^{K}\tilde{N}_{k}$, and $\mathcal{\tilde{S}}^{(k)} = \{ G_{1}^{(k)}, \ldots, G_{\tilde{N}_{k}}^{(k)} \}$, $k=1, \ldots, K$, is denoted as a set of graphs allocated with label $y_{k}$. The test set $\mathcal{G}^{Test} \subseteq \mathcal{G}$ is next sampled by rate $\zeta$.

Next, we represent each $G_{i} \in \mathcal{G}^{Train} \cup \mathcal{G}^{Test}$ by stacking all its distances to the support set in a $1 \times \tilde{N}$ vector:
\begin{equation}\label{graph-MetricDistribution2vec}
  \tilde{M}_{G_{i}} = [\frac{d_{i,1}^{(1)}}{\sigma_{i}}, \ldots, \frac{d_{i, \tilde{N}_{1}}^{(1)}}{\tilde{\sigma}_{i}}, \ldots, \frac{d_{i,1}^{(K)}}{\tilde{\sigma}_{i}}, \ldots, \frac{d_{i, \tilde{N}_{K}}^{(K)}}{\tilde{\sigma}_{i}}]_{1 \times \tilde{N}},
\end{equation}
where $\tilde{\sigma}_{i} = \sum_{k = 1}^{K} \sum_{j = 1}^{\tilde{N}_{k}} d_{i,j}^{(k)}$. We define $\tilde{M}_{G_{i}}$ as the embedding of $G_{i}$.

Note that, we address two remarks for the selection of training set $\mathcal{G}^{Train}$ since the dataset $\mathcal{G}$ may suffer a data imbalance issue.
\begin{itemize}
\item The training set should cover all the categories contained in $\mathcal{G}$.
\item On the premise of the first remark, the number of different labeled instances is excepted to coincide with its proportion in the dataset.
\end{itemize}

\begin{algorithm}
	\renewcommand{\algorithmicrequire}{\textbf{Input:}}
	\renewcommand{\algorithmicensure}{\textbf{Output:}}
	\caption{Metric distribution to vector}
	\label{Algorithm-Metric-distribution-to-vector}
	\begin{algorithmic}[1]
		\REQUIRE graph dataset $\mathcal{G} = \{G_{i}, G_{2}, \ldots, G_{N}\}$, sampling rate $\eta$ for training set, sampling rate $\zeta$ for test set, and metric function $d$.
		\ENSURE embedding for each training graph and test graph.

        \STATE evenly sample $\tilde{N}$ training items in training set $\mathcal{G}^{Train}$ by sampling rate $\eta$ and a test set $\mathcal{G}^{Test}$
        \STATE split $\mathcal{G}^{Train}$ into $K$ disjoint equivalence sets $\mathcal{G}^{Train} = \bigcup_{k = 1}^{K} \mathcal{\tilde{S}}^{(k)}$, where $\mathcal{\tilde{S}}^{(k)} = \{ G_{1}^{(k)}, \ldots, G_{\tilde{N}_{k}}^{(k)} \}$
        \STATE $ \mathcal{G}^{Train} = \{G_{1}^{(1)}, \ldots, G_{\tilde{N}_{1}}^{(1)}, \ldots , G_{1}^{(K)}, \ldots, G_{\tilde{N}_{K}}^{(K)} \}$

        \FOR {each $G_{i} \in \mathcal{G}^{Train} \cup \mathcal{G}^{Test}$}
            \FOR {$k=1$; $k \leq K$; $k++$}
                \FOR {$j=1$; $j \leq \tilde{N}_{k}$; $j++$}
                    \STATE $d_{i,j}^{(k)} = d(G_{i}, G_{j}^{(k)})$
                \ENDFOR
            \ENDFOR
            \STATE $\tilde{\sigma}_{i} = \sum_{k = 1}^{K} \sum_{j = 1}^{\tilde{N}_{k}} d_{i,j}^{(k)}$
            \STATE $\tilde{M}_{G_{i}} = [\frac{d_{i,1}^{(1)}}{\tilde{\sigma}_{i}}, \ldots, \frac{d_{i, \tilde{N}_{1}}^{(1)}}{\tilde{\sigma}_{i}}, \ldots, \frac{d_{i,1}^{(K)}}{\tilde{\sigma}_{i}}, \ldots, \frac{d_{i, \tilde{N}_{K}}^{(K)}}{\tilde{\sigma}_{i}}]_{1 \times \tilde{N}}$
        \ENDFOR

        \STATE \textbf{return} $\tilde{M}_{G_{i}}$, where $G_{i} \in \mathcal{G}^{Train} \cup \mathcal{G}^{Test}$
	\end{algorithmic}
\end{algorithm}

The striking difference apparent from the pioneering embedding works is that our representative principle reveals the characteristics of every single graph in the view of metric distribution In contrast, the previous representative methods dedicate to various vectorial representations for each graph in terms of topology, features, etc. Or in short, we tend to embed each graph entity by its broad-scale discriminations in similarities or discrepancies instead of an accurate depiction of various inner properties. This is the fundamental opinion of this representation strategy.

\begin{figure}[h]
\centering
\includegraphics[height=6cm,width=8cm]{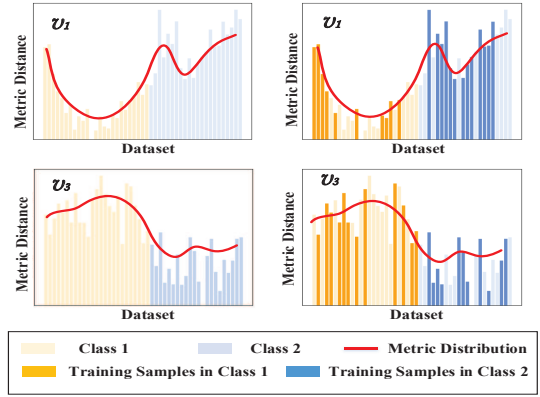}
\caption{The top right-hand and bottom right-hand figures are plotted by 50\% of the dataset set as the training set, marked in dark colors. Their overall trends of metric distribution still roughly hold as the top left-hand and bottom left-hand figures, respectively.
}
\label{Figure-Metric-Distribution-Selected}
\end{figure}

The apparent advantage of representing structural data with metric distribution is that distribution-based embedding could eliminate the reliance on abundant training samples. The formula~(\ref{graph-MetricDistribution2vec}), differing very much from (\ref{norm-metric-distribution}), depends only on the pre-selected training samples and the targeted graph to be embedded, rather than on the entire dataset $\mathcal{G}$. Correspondingly, the overall computational complexity is thus reduced to $\emph{O}[N \cdot \tilde{N} \cdot \emph{O}(d)]$.

As we separate $G$ into an union of $K$ disjoint segmentations annotated with different labels, i.e.,
\begin{equation}
  \mathcal{G} = \bigcup\limits_{k = 1}^{K} \mathcal{S}^{(k)},
\end{equation}
accordingly, the selected training samples in $\mathcal{\tilde{S}}^{(k)}$ are employed to stretch its corresponding $\mathcal{S}^{(k)}$ as much as possible. We train the classifier $f$ on the training set $\mathcal{G}^{Train}$, and predict the classification outcome at each test $G$ according to its global distribution trend within the training set $\mathcal{G}^{Train}$ scale, which is used to represent the entire database. However, the volume of the training set simply accounts for the approximation of the ground-truth metric distribution in general, but matters little to the overall trend of metric distribution at the targeted point, shown as the red lines in Figure~\ref{Figure-Metric-Distribution-Selected}. This fact will provide practicable application on Few-Shot classification, where only limited training samples are used for training. More experimental validations of effectiveness in Few-Shot classification could be seen in Section~\ref{Experiments-Few-Shot-Classification}.

\section{Distance metric induced by optimal transportation}\label{optimal-transportation}
Choosing an appropriate metric distance learned from the graph-structured data is vital to the distance-based algorithms in the area of graph pattern classification. In this part, we introduce a novel graph metric distance design using the classic optimal transportation~\cite{optimal_transportation_villani}, which measures the transportation effort between graph generation distributions as distance. We start with the mathematical preliminaries associated with optimal transportation. Then, we will build the framework of optimal transportation between graph data in a discrete manner.

\subsection{Wasserstein distance of optimal transportation}\label{section-Wasserstein-distance}
The optimal transportation due to Monge and Kantorovich is used to measure the distance between probability distributions defined on a given
metric space. Monge first introduced the optimal transference mapping, and Kantorovich extended optimal transference mapping to optimal transference plan and proposed the equivalent dual problem of optimal transportation~\cite{optimal_transportation_villani}. It comes from the following realistic scenes: given a pile of sand and a hole that we have to completely fill up with the sand, how to transport the sand with the minimum cost if we assume there need efforts to move the sand to the hole. If we normalize the mass of the pile to 1, then we model both the pile and the hole by probability measures $\mu \in \mathcal{P}(X)$ and $\nu \in \mathcal{P}(Y)$, where $X$, $Y$ denote two metric spaces and $\mathcal{P}(X)$, $\mathcal{P}(Y)$ are the corresponding probability spaces.

The \textbf{Monge-Kantorovich distance} $W_{p}$~\cite{Monge-Kantorovich-distance} is referred to quantify the minimum expected Kantorovich's optimal transportation cost between $X$ and $Y$ endowed with distance $d$, and is defined as
\begin{equation}\label{Wp-cost-function-discrete}
  W_{p}(X, Y) = [\inf \limits_{\pi \in \Pi(\mu, \nu)} \int_{X \times Y}d^{p}(x, y)d\pi(x, y)] ^{\frac{1}{p}},
\end{equation}
where $\Pi(\mu, \nu)$ contains a set of transference plans $\pi$s which satisfy marginal distribution condition (MD),
\begin{equation}\label{transference-plan-set}
  \Pi(\mu, \nu) = \{\pi \in \mathcal{P}(X \times Y)|\pi \; holds \;\mathrm{MD}\}.
\end{equation}

In particular, $W_{2}(G_{i}, G_{j})$ will be referred as the \textbf{Wasserstein distance}~\cite{fused_Gromov_Wasserstein_Distances} between $X$ and $Y$ when exponent $p = 2$,
\begin{equation}\label{Wasserstein-cost-function}
  W_{2}(X, Y) = [\inf \limits_{\pi \in \Pi(\mu, \nu)} \int_{X \times Y}d^{2}(x, y)d\pi(x, y)] ^{\frac{1}{2}}.
\end{equation}

\subsection{Optimal transportation on graphs}\label{Optimal-Transportation-between-Graphs}
Following the above conceptions, we may want to introduce the framework of optimal transportation to the distance measured between different structural graph data. Our intuition is that when we take each data as an individual among the probability spaces for generating graphs, the structural graphs tend to share proximity if they suffer fewer costs once they conduct optimal transportation between pairwise distributions.

\subsubsection{Graph frequent fragment decomposition}
The first thing we need to make clear is the choice of ``sand'' in this graph transportation scenario, i.e., to establish the pre-images and images in optimal transportation.
In structural graph data, nodes are always clustered as mesoscopic substructures to express physical, chemical, or socially interactive functions. Thus, we take these functional units as the materials transferred between pairwise graphs. A substructure $sg$ is denoted as a \textbf{fragment} of graph $G$ from graph dataset $\mathcal{G} = \{G_{1}, \ldots, G_{N}\}$ if its nodes satisfy $\mathcal{V}(sg) \subseteq \mathcal{V}(G)$, and edges satisfy $\mathcal{E}(sg) \subseteq \mathcal{E}(G)$. We assume $\mathcal{SG}$ as the set that contains all possible fragments of graph dataset $\mathcal{G}$.

\begin{figure}[h]
\centering
\includegraphics[height=6cm,width=7cm]{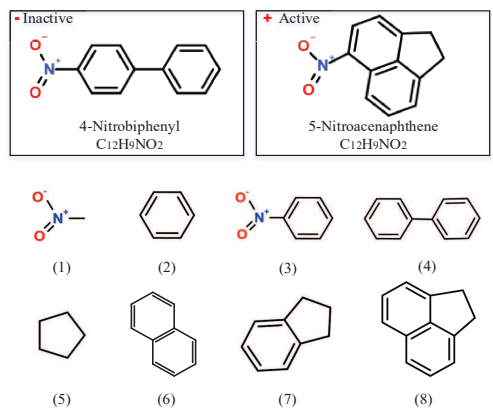}
\caption{An illustration of the \emph{frequent fragment} and \emph{min-sup} threshold $\theta$.
Chemically, 4-Nitrobiphenyl and 5-Nitroacenaphthene are isomers with the same molecular formula $\textrm{C}_{12}\textrm{H}_{9}\textrm{NO}_{2}$ while they possess different properties in medical applications. If $\theta = 0.5$, all the eight subgraphs could be regarded as frequent fragments. When $\theta = 1$, only subgraphs (1), (2), and (3) are within the scope of frequent fragments.
}
\label{Figure-FSG}
\end{figure}

When discriminating two targeted graphs, an intuition tells that they are analogous in topology if they share similar fragment decomposition. However, it shall be unrealistic to compare all possible graph fragments due to the inevitable complexity of calculation. More to the point, not each fragment contributes equally, and only a fraction of fragments are relevant to the graph representation. The frequent fragments of a graph dataset denote the substructures (including subsequences or subgraphs) that appear in a graph dataset with a frequency no less than a pre-specified threshold, $\emph{min-sup}$. In specifically, a fragment $sg \in \mathcal{SG}$ shall be called a \textbf{frequent fragment} $fsg$ if $sg$ satisfies
\begin{equation}\label{support}
  Supp = \frac{|\{G_{i} \in \mathcal{G} |sg \in G_{i}\}|}{|\mathcal{G}|} \geq \theta,
\end{equation}
where $\theta \in [0, 1]$ is the $\textbf{\emph{min-sup}}$ threshold. A concrete example to illustrate the frequent fragment and the $\emph{min-sup}$ is stated in Figure~\ref{Figure-FSG}.

Frequent fragment pattern, as a form of non-linear feature combinations over the set of different substructures, has higher discriminative power than a single kind of fragment because they capture more inner associations and underlying semantics of the data. Thus finding frequent fragments plays a focused theme in data mining research, and the key point is to specify the hyper-parameter $\emph{min-sup}$ threshold $\theta$ used in the model in a frequent pattern. Once the $\emph{min-sup}$ threshold $\theta$ is specified, we define the \textbf{frequent fragment decomposition} of graph $G_{i}$ by $D(G_{i})$, as the set of frequent fragments that hold inequality (\ref{support}). The union of each frequent fragment decomposition $D(G_{i})$ for graph set $\mathcal{G}$ is given by
\begin{equation}
  D(\mathcal{G}) = \bigcup \limits_{i=1}^{N} D(G_{i}).
\end{equation}

\subsubsection{Embedded frequent fragment}
In actual pattern recognition tasks, it is not feasible to directly apply the frequent graphical fragments to the expression of optimal transportation. Recent works of embedding methods inspire us to represent each frequent fragment by a $d$-dimension vector in the Euclidean space. Thus here, we adopt the PV-DBOW~\cite{PV-DBOW} technique to realize the conversion of data type from the topological form into vectorial code.

PV-DBOW is commonly used in Natural Language Processing to learn an arbitrary size representation of a document from the document set as well as the words contained in the document. Now if we regard the given graph set $\mathcal{G} = \{G_{1}, \ldots, G_{N}\}$ as the document set, and the frequent component decomposition $D(G_{i})$ as the word set for targeted $G_{i}$, the PV-DBOW tends to maximize the logarithmic probability of predicting its frequent fragments under the condition of $G_{i}$:
\begin{equation}\label{log-probability}
  \sum\limits_{fsg \in D(G_{i})} \log \textmd{Pr}(fsg | G_{i}).
\end{equation}
The conditional probability $\textmd{Pr}(fsg|G_{i})$ is defined as a softmax unit parameterized by a dot product
\begin{equation}\label{conditional-probability}
  \textmd{Pr}(fsg | G_{i})=
  \frac{\exp(\boldsymbol{fsg} \cdot \boldsymbol{G_{i}})}
  {\sum\limits_{fsg' \in D(\mathcal{G})}\exp(\boldsymbol{fsg'} \cdot \boldsymbol{G_{i}})},
\end{equation}
where $\boldsymbol{G_{i}}$ and $\boldsymbol{fsg}$ denote the $dim$-dimension vectorial representations of $G_{i}$ and $fsg$.

We collect all the embedding of frequent fragments from graph $G_{i}$ in set $E(G_{i})$. The union of all the $E(G_{i})$s is given by
\begin{equation}
  E(\mathcal{G}) = \bigcup \limits_{i=1}^{N} E(G_{i}).
\end{equation}

\subsubsection{Wasserstein distance on graphs}
\begin{figure*}[h]
\centering
\includegraphics[height=6cm,width=16cm]{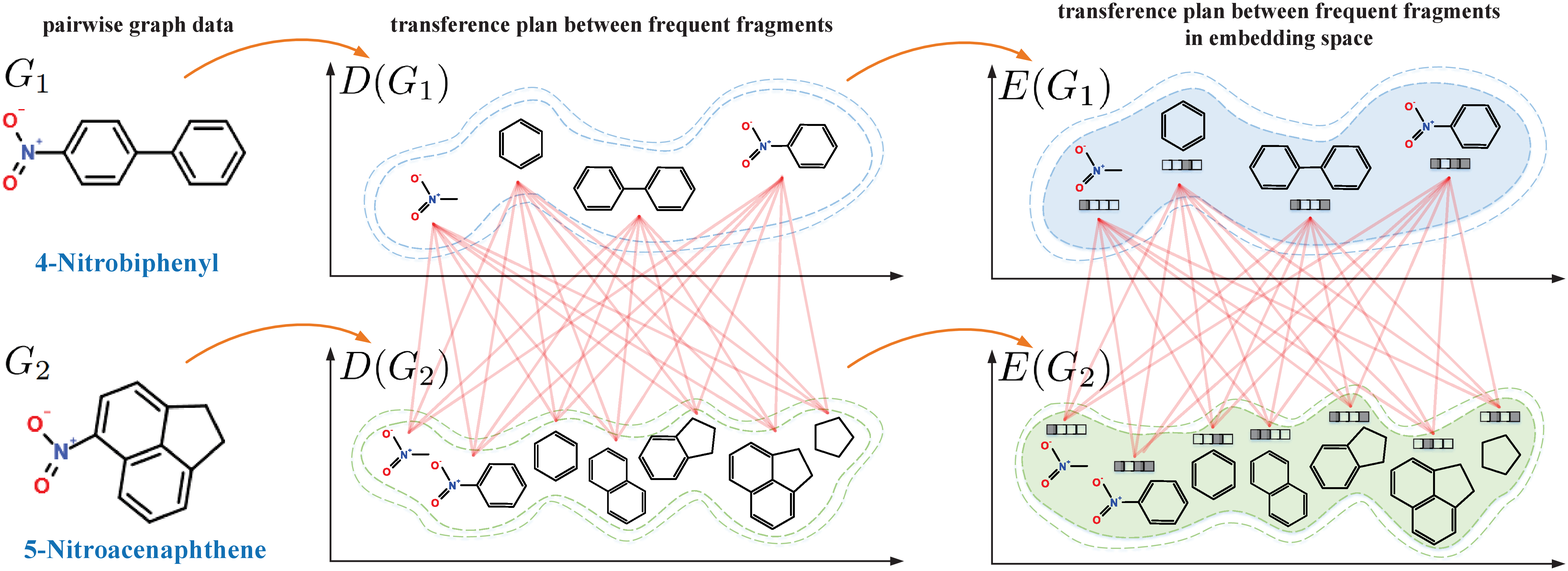}
\caption{The illustration of the optimal transportation between frequent fragment decompositions and between vectorial frequent fragment decompositions. The cluster of red lines denotes the transference plan for this transportation scenario.
}
\label{Figure-KantorovichOT-FSG}
\end{figure*}

Based on the above definitions, the Wasserstein distance (\ref{Wasserstein-cost-function}) on graphs could be written in a discrete manner as
\begin{equation}\label{discrete-quadratic-Wasserstein-distance}
\begin{aligned}
  W_{2}(G_{i}, G_{j})   & = [\inf \limits_{\pi \in \Pi(\tilde{\mu}, \tilde{\nu})} \sum_{fsg \in D(G_{i})} \sum_{fsg' \in D(G_{j})} d^{2}(fsg, fsg')] ^{\frac{1}{2}}\\
                        & = [\inf \limits_{\pi \in \Pi(\tilde{\mu}, \tilde{\nu})} \sum_{\boldsymbol{fsg} \in E(G_{i})} \sum_{\boldsymbol{fsg'} \in E(G_{j})} d^{2}(\boldsymbol{fsg}, \boldsymbol{fsg'})] ^{\frac{1}{2}},
\end{aligned}
\end{equation}
where $\Pi(\tilde{\mu}, \tilde{\nu}) = \{\pi \in \mathcal{P}[D(G_{i}) \times D(G_{j})]| \pi~holds~\mathrm{MD}\}$ is the transference plan set on $\tilde{\mu} \in \mathcal{P}[D(G_{i})]$ and $\tilde{\nu} \in \mathcal{P}[D(G_{j})]$, and $d^{2}(\boldsymbol{fsg}, \boldsymbol{fsg'})$ denotes the cost of mapping $\boldsymbol{fsg} \in E(G_{i})$ to $\boldsymbol{fsg'} \in E(G_{j})$ by Euclidean distance. The optimal transportation framework is shown in Figure~\ref{Figure-KantorovichOT-FSG}.

It has proved that the solution to optimal transportation problems faces a heavy computational price tag, because the computing cost scales at least $O(k^{3}log(k))$~\cite{comp-complex-OT}, where $k$ is the least volume of frequent fragment decomposition for each graph in the dataset. To speed up the calculation, Sinkhorn~\cite{sinkhorn2021} tackles the original transportation problems with an entropic regularization term and then turns the transportation problem into a strictly convex problem that can achieve a linear convergence by matrix scaling procedures. Thus in this paper, we leverage the classical Sinkhorn Algorithm (parameterized by the regularization coefficient $\lambda$ and the maximum iteration times $t$) to find the optimal transference and derive the minimum quadratic Wasserstein distance. More details about the Sinkhorn Algorithm are stated in the works~\cite{sinkhorn2021, sinkhorn2020}.

\section{Experiments}\label{Experiments}
In this section, we report the comprehensive experiments to validate the efficiency of our proposed methodology on both supervised classification and Few-Shot classification problems. We focus on the supervised pattern tasks with training data and desired output labels. It shows our method could achieve superior performance compared with several well-used baselines.

\subsection{Standard Datasets}
Our proposed method is tested on a series of real-world graph datasets: \textbf{DD}, \textbf{MUTAG}, \textbf{PTC}, \textbf{PROTEINS}, \textbf{ENZYMES}, \textbf{IMDB-Multi}, \textbf{Tox21-AR}, \textbf{Tox21-ER}, \textbf{NCI-1}, \textbf{NCI-33}, \textbf{NCI-83}, and \textbf{NCI-109}. The brief descriptions are as follows and their statistics are summarized in Table \ref{Table-dataset-stats}.

\begin{table}[htb]
\centering
\caption{Statistics of the benchmark graph datasets. The columns are the name of the dataset, the number of graphs, the number of classes, the average number of nodes, and the average number of edges. Here we mention that each dataset is balanced in the number of different labeled parts, and the NCI-1, NCI-33, NCI-83, and NCI-109 are all sampled randomly from the original datasets as their vast volumes.}
\begin{tabular}{lcccc}
\toprule
\toprule
\textbf{Dataset}        & \textbf{Total}    & \textbf{Class}    & \textbf{Average Nodes} & \textbf{Average Edges} \\
\midrule
\textbf{DD}             & 1178              & 2                 & 284.32                & 715.66\\
\textbf{MUTAG}          & 188               & 2                 & 17.93                 & 19.79\\
\textbf{PTC}            & 344               & 2                 & 14.29                 & 14.69\\
\textbf{PROTEINS}       & 1112              & 2                 & 39.06                 & 72.82\\
\textbf{ENZYMES}        & 600               & 6                 & 32.63                 & 62.13\\
%\textbf{IMDB-Binary}    & 1000              & 2                 & 19.78                 & 96.39\\
\textbf{IMDB-Multi}     & 1500              & 3                 & 13.00                 & 65.94\\
\textbf{Tox21-AR}       & 760               & 2                 & 27.09                 & 28.83\\
\textbf{Tox21-ER}       & 1874              & 2                 & 22.79                 & 24.02\\
\textbf{NCI-1}          & 800               & 2                 & 29.87                 & 32.30\\
\textbf{NCI-33}         & 800               & 2                 & 32.69                 & 35.48\\
\textbf{NCI-83}         & 800               & 2                 & 33.71                 & 36.26\\
\textbf{NCI-109}        & 800               & 2                 & 29.68                 & 32.13\\
\bottomrule
\end{tabular}
\label{Table-dataset-stats}
\end{table}

\begin{itemize}
    \item \textbf{DD}~\cite{dataset-dd}: DD is a dataset of protein structures where nodes represent amino acids and edges indicate spatial closeness, classiﬁed into enzymes or non-enzymes.
    \item \textbf{MUTAG}~\cite{dataset-mutag}: MUTAG is a dataset of aromatic and heteroaromatic nitro compounds labeled according to whether they have a mutagenic effect on bacteria or not.
    \item \textbf{PTC}~\cite{dataset-ptc-mr}: PTC consists of graph representations of chemical molecules labeled according to carcinogenicity for male and female rats.
    \item \textbf{PROTEINS}~\cite{dataset-protein}: PROTEINS is a set of protein graphs where nodes represent secondary structure elements and edges indicate neighborhood in the amino-acid sequence 3-dimension space.
    \item \textbf{ENZYMES}~\cite{dataset-protein}: ENZYMES consists of protein tertiary structures obtained from the BRENDA enzymes datasets.
    %\item \textbf{IMDB-Binary}~\cite{dataset-imdb-binary}: IMDB-Binary is a dataset built from the Internet Movie Data (IMDB). Each data from IMDB-Binary will be divided into positive or negative according to the comments on movies.
    \item \textbf{IMDB-Multi}~\cite{dataset-imdb-binary}: IMDB-Multi is a dataset built from the Internet Movie Data (IMDB). The data from IMDB-Multi will be divided into three categories.
    \item \textbf{Tox21}~\cite{dataset-Tox21}: Tox21 is from a federal collaboration program ``Toxicology in the 21st Century (Tox21)''. Tox21 aims to develop better toxicity assessment methods to quickly and efficiently test whether certain chemical compounds can potentially disrupt human body processes that may lead to negative health effects. Among these datasets, Tox21-AR is used to test toxicity for the androgen receptors, and Tox21-ER is to test toxicity for estrogen receptors.
    \item \textbf{NCI}~\cite{dataset-nci}: NCI is a series of chemical compound datasets, which are divided by the anti-cancer properties (active or negative). The National Cancer Institute (NCI) has made these datasets publicly available. The NCI-1, NCI-33, NCI-83, and NCI-109 datasets contain molecules labeled by medical applications for Non-Small Cell Lung Cancer, Melanoma Cancer, Breast Cancer, and Ovarian Cancer respectively.
\end{itemize}

\subsection{Compared Baselines}
To fully illustrate the notable performance of our model, we explore a series of baselines that are summarized as follows.

\begin{itemize}
    \item \textbf{Weisfeiler-Lehman}~\cite{WL-kernel}: Weisfeiler-Lehman kernel maps graph data into a Weisfeiler-Lehman sequence, whose node attributes represent graph topology and label information. Weisfeiler-Lehman kernel is wildly used in isomorphism tests on graphs since the runtime scales linearly in the number of edges of the graphs and the length of the Weisfeiler-Lehman graph sequence.
    \item \textbf{Graphlet}~\cite{GK-graphlet-kernel}: Graphlet kernel measures graph similarity by counting common $k$-node graphlets, ensuring the computation complexity is restricted in polynomial time.
    \item \textbf{Deep Weisfeiler-Lehman, Deep Graphlet}~\cite{Deep-graph-kernels}: Deep Weisfeiler-Lehman kernel and Deep Graphlet kernel leverage language modeling and deep learning to learn latent representations of sub-structures for graphs.
    \item \textbf{Persistence Fisher}~\cite{Persistence-Fisher}: Persistence Fisher kernel relies on Fisher information geometry to explore persistence diagrams on structural graph pattern recognition.
    \item \textbf{WKPI}~\cite{WKPI}: WKPI designs a weighted kernel for persistence images, together with a metric-learning method to learn the best classify function for labeled data.
    \item \textbf{Graph2vec}~\cite{graph2vec}: Graph2vec treats rooted subgraphs as words and graphs as sentences or documents, and then it uses Skip-gram in Natural Language Processing to get explicit graph embeddings.
    \item \textbf{GE-FSG}~\cite{GE-FSG}: GE-FSG decomposes each graph data into a set of sub-structures and then maps the graph into embedding space using a PV-DBOW technique. GE-FSG discriminates two graphs by the similarity of their embedding vectors.
    \item \textbf{AWE}~\cite{AWE}: AWE uses anonymous random walks to embed entire graphs in an unsupervised manner, but it takes a different embedding strategy compared with our methodology. AWE leverages the neighborhoods of anonymous walks while our work focuses on the co-occurring anonymous walks on a global scale.
    \item \textbf{PATCHY-SAN}~\cite{PSCN}: Analogous to convolutional neural networks, PATCHY-SAN proposes a framework to perform convolutional operations for arbitrary graph data.
    \item \textbf{GCN}~\cite{GCN}: GCN (Graph Convolutional Networks) uses Laplacian multiplication to construct a convolutional operator suitable for structural graph data.
\end{itemize}

\subsection{Implementation}
This paper uses Python 3.7.9, Pytorch 1.0.2, Numpy 1.22.0rc2, Scikit-learn 0.24.1, POT 0.8.1.0, and Gensim 3.8.3 as the computing environment. All experiments conducted on the workstation with 2 INTEL XEON CPUs and 4 NVIDIA GeForce GTX1080Ti GPUs. The parallel computational technique is adopted to accelerate the calculation, where the procedure is divided into 20 parallel sub-processes. In addition, the ten-fold cross-validation is used to estimate the experimental result of each dataset. To perform classification, we fit three lightweight classifiers, i.e., kNN, Logistic Regression, and SVM with RBF kernel using the training set, and validate the classification on the test set. This procedure is performed ten times and the average accuracy on the targeted dataset is considered as an evaluation of our method. In the last, we fix the vital parameters involved in Word2Vec, Sinkhorn, kNN, and SVM (RBF Kernel) in advance, as shown in Table~\ref{Table-Parameters}.

\begin{table*}[htb]
\centering
\caption{The main parameters used in Word2Vec, Sinkhorn, kNN, and SVM (RBF Kernel).}
\begin{tabular}{lccc}
\toprule
\toprule
\textbf{Algorithm}          &\textbf{Parameter}     & \textbf{Value}   & \textbf{Explanation}    \\
\midrule
\textbf{Word2Vec}           & $dim$                 & 16                &Embedding dimension of each frequent fragment  \\
\textbf{Sinkhorn}           & $\lambda$             & $10^{-2}$         &Regularization coefficient         \\
\textbf{Sinkhorn}           & $t$                   & 30                &Maximum iteration time        \\
\textbf{kNN}                & $k$                   & 3                 &Number of neighbors        \\
\textbf{SVM (RBF kernel)}   & $C$                   & $[10^{-6}, 10^{-4}, 10^{-2}, 1, 10^{2}, 10^{4}, 10^{6}]$  &Regularization coefficient selected by cross-validation(CV)\\
\bottomrule
\end{tabular}
\label{Table-Parameters}
\end{table*}

\subsection{Parameter Selection}
In this paper, we mainly concentrate on the \emph{min-sup} threshold, which is relevant to the selection of frequent fragments from the entire graph set. To ascertain the \emph{min-sup} that induces the best experimental result, we traverse the values of \emph{min-sup} from large to small and then report the accuracies of classifiers including kNN, Logistic Regression, and SVM with RBF kernel as well as the number of selected frequent fragments on 12 datasets in the point-fold line charts in Figure~\ref{Figure-Sensitivity}. We take the highest points of each overall performance as the selected \emph{min-sup} for MetricDistribution2vec using kNN, Logistic Regression, or SVM (RBF Kernel) as a classifier.

\subsection{Experimental Results and Discussion}
\begin{table*}[htb]
\centering
\caption{Summary of the average classification accuracy (standard deviation) (\%) for baselines. The result of each dataset is over the 10-fold stratified cross-validation and is under the same random seed setting. ``--'' means the average classification accuracy (standard deviation) is unavailable from the published papers. The experimental results of MetricDistribution2vec using kNN, Logistic Regression, or SVM (RBF Kernel) under a best-selected \emph{min-sup} are also reported. For each dataset, the best baseline is marked in blue and the best accuracies of MetricDistribution2vec with different classifiers are observed in orange, green, and red, respectively.
}
\begin{tabular}{|l||cccccc|}
%\toprule
\hline
\textbf{Algorithm}                              & \textbf{DD}       & \textbf{MUTAG}    & \textbf{PTC}     & \textbf{PROTEINS}	& \textbf{ENZYMES}  & \textbf{IMDB-Multi}  \\
\hline
Weisfeiler-Lehman~\cite{WL-kernel}              & 77.95 (0.70)      & 80.63 (3.07)      & 56.97 (2.01)      & 72.92 (0.56)& 53.15 (1.14)        & 50.55 (0.55)\\
Deep Weisfeiler-Lehman~\cite{Deep-graph-kernels}& --                & 82.94 (2.68)      & 59.17 (1.56)      & 73.30 (0.82)& \textcolor{blue}{\textbf{53.43 (0.91)}}  	& --\\
Graphlet~\cite{GK-graphlet-kernel}              & 78.45 (0.26)      & 81.66 (2.11)      & 57.26 (1.41)      & 71.67 (0.55)& 26.61 (0.99)        & 43.89 (0.38)\\
Deep Graphlet~\cite{Deep-graph-kernels}         & 79.10 (2.50)      & 82.66 (1.45)      & 66.96 (0.56)      & 71.68 (0.50)& 27.08 (0.79)        &44.55 (0.52)\\
Persistence Fisher~\cite{Persistence-Fisher}    & 79.40 (0.80)      & 85.60 (1.70)      & 62.42 (1.80)      & 75.20 (2.10)& --                  & 48.60 (0.70) \\
WKPI~\cite{WKPI}                                & 82.00 (0.50)      & 85.80 (2.50)      & 62.70 (2.70)      & 78.50 (0.40)& --                  & 49.50 (0.40) \\
Graph2Vec~\cite{graph2vec}                      & 58.64 (0.01)      & 83.15 (9.25)      & 60.17 (6.86)      & 73.30 (2.05)& 44.33 (0.09)        & 45.47 (0.04)\\
GE-FSG~\cite{GE-FSG}                            &\textcolor{blue}{\textbf{91.69 (0.02)}}          &71.00(2.29)    &\textcolor{blue}{\textbf{73.00 (0.04)}}                 & \textcolor{blue}{\textbf{81.79 (0.04)}}&84.74 (0.07)       &\textcolor{blue}{\textbf{55.22 (0.05)}}       \\
AWE~\cite{AWE}                                  & 71.51 (4.02)      & 87.87 (9.76) & 59.14 (1.83) & 70.01 (2.52) & 35.77 (5.93) &51.58 (4.66) \\
PATCHY-SAN~\cite{PSCN}                          & 77.12 (2.41)      &\textcolor{blue}{\textbf{92.63 (4.21)}}  & 60.00 (4.82)      & 75.89 (2.76)& --             & 45.23 (2.84)        \\
GCN~\cite{GCN}                                  & 66.83 (4.30)      & 91.64 (7.20)       & 71.35 (0.64)      & 67.21 (3.00)& 24.26 (4.70)      & --         \\
\hline
\textbf{MetricDistribution2vec + kNN} & \textcolor[rgb]{ .749,  .561,  0}{\textbf{98.22 (0.80)}} & \textcolor[rgb]{ .749,  .561,  0}{\textbf{98.42 (2.42)}}  & \textcolor[rgb]{ .749,  .561,  0}{\textbf{92.71 (4.00)}} & \textcolor[rgb]{ .749,  .561,  0}{\textbf{98.83 (1.06)}} & \textcolor[rgb]{ .749,  .561,  0}{\textbf{86.84 (3.20)}} & \textcolor[rgb]{ .749,  .561,  0}{\textbf{96.67 (1.76)}}\\
\textbf{MetricDistribution2vec + Logistic Regression} & \textcolor[rgb]{ .329,  .51,  .208}{\textbf{98.65 (0.86)}} & \textcolor[rgb]{ .329,  .51,  .208}{\textbf{95.20 (3.78)}}  & \textcolor[rgb]{ .329,  .51,  .208}{\textbf{94.76 (2.57)}} & \textcolor[rgb]{ .329,  .51,  .208}{\textbf{98.03 (1.12)}} & \textcolor[rgb]{ .329,  .51,  .208}{\textbf{92.17 (2.79)}} & \textcolor[rgb]{ .329,  .51,  .208}{\textbf{77.94 (4.04)}}\\
\textbf{MetricDistribution2vec + SVM (RBF Kernel)} & \textcolor[rgb]{ 1,  0,  0}{\textbf{99.41 (0.76)}} & \textcolor[rgb]{ 1,  0,  0}{\textbf{97.86 (2.62)}}  & \textcolor[rgb]{ 1,  0,  0}{\textbf{97.66 (2.88)}} & \textcolor[rgb]{ 1,  0,  0}{\textbf{99.19 (0.63)}} & \textcolor[rgb]{ 1,  0,  0}{\textbf{94.50 (2.99)}} & \textcolor[rgb]{ 1,  0,  0}{\textbf{81.80 (3.20)}}\\
\hline
\emph{min-sup}                                  & 0.85              & 0.95              & 0.45              & 0.75      & 0.85                  & 0.30 \\
\hline

\hline
\hline
\textbf{Algorithm}                              & \textbf{Tox21-AR} & \textbf{Tox21-ER} & \textbf{NCI-1}    & \textbf{NCI-33}       & \textbf{NCI-83}   & \textbf{NCI-109}\\
\hline

Weisfeiler-Lehman~\cite{WL-kernel}              & 69.23 (1.03)      & 65.81 (2.32)      & 80.13 (0.50)      & 78.98 (0.36)          & 79.54 (0.26)      & 80.22 (0.34)\\
Deep Weisfeiler-Lehman~\cite{Deep-graph-kernels}& 71.05 (0.98)      & 66.42 (1.95)      & 80.31 (0.46)      & 81.24 (0.76)          & 80.29 (0.46)      & 80.32 (0.33)\\
Graphlet~\cite{GK-graphlet-kernel}              & 74.00 (0.83)      & 71.26 (0.66)      & 62.28 (0.29)      & 61.78 (0.31)          & 63.02 (0.18)      & 62.60 (0.19)\\
Deep Graphlet~\cite{Deep-graph-kernels}         & 74.02 (0.24)      & 72.68 (0.34)      & 62.48 (0.25)      & 62.02 (0.35)          & 63.88 (0.19)      & 62.48 (0.25)\\
GE-FSG~\cite{GE-FSG} 							&\textcolor{blue}{\textbf{79.53 (0.55)}}&\textcolor{blue}{\textbf{77.56 (0.54)}}&\textcolor{blue}{\textbf{84.36 (0.02)}}&\textcolor{blue}{\textbf{84.99 (0.11)}}&\textcolor{blue}{\textbf{83.34 (0.08)}}&\textcolor{blue}{\textbf{85.59 (0.01)}}\\
AWE~\cite{AWE}                                  & 72.74 (3.98)      & 71.51 (3.73)      & 65.77 (5.93)      & 68.54 (2.99)          & 70.22 (3.23)      & 70.01 (2.52)\\
GCN~\cite{GCN}                                  & 76.13 (0.32)      & 72.35 (0.64)      & 76.27 (4.10)      & 77.49 (3.82)          & 79.01 (2.64)      & 80.47 (0.19)\\
\hline
\textbf{MetricDistribution2vec + kNN} & \textcolor[rgb]{ .749,  .561,  0}{\textbf{96.45 (2.20)}} & \textcolor[rgb]{ .749,  .561,  0}{\textbf{98.67 (0.72)}} & \textcolor[rgb]{ .749,  .561,  0}{\textbf{95.38 (1.77)}} & \textcolor[rgb]{ .749,  .561,  0}{\textbf{94.25 (2.45)}} & \textcolor[rgb]{ .749,  .561,  0}{\textbf{93.50 (2.29)}} & \textcolor[rgb]{ .749,  .561,  0}{\textbf{93.50 (2.95)}} \\
\textbf{MetricDistribution2vec + Logistic Regression} & \textcolor[rgb]{ .329,  .51,  .208}{\textbf{98.16 (1.34)}} & \textcolor[rgb]{ .329,  .51,  .208}{\textbf{98.67 (1.10)}} & \textcolor[rgb]{ .329,  .51,  .208}{\textbf{99.63 (0.57)}} & \textcolor[rgb]{ .329,  .51,  .208}{\textbf{98.88 (1.42)}} & \textcolor[rgb]{ .329,  .51,  .208}{\textbf{98.75 (1.12)}} & \textcolor[rgb]{ .329,  .51,  .208}{\textbf{99.25 (0.83)}} \\
\textbf{MetricDistribution2vec + SVM (RBF Kernel)} & \textcolor[rgb]{ 1,  0,  0}{\textbf{97.63 (1.64)}} & \textcolor[rgb]{ 1,  0,  0}{\textbf{99.36 (0.52)}} & \textcolor[rgb]{ 1,  0,  0}{\textbf{99.38 (0.62)}} & \textcolor[rgb]{ 1,  0,  0}{\textbf{99.75 (0.75)}} & \textcolor[rgb]{ 1,  0,  0}{\textbf{99.00 (1.09)}} & \textcolor[rgb]{ 1,  0,  0}{\textbf{99.13 (0.57)}} \\
\hline
\emph{min-sup}                            & 0.55              & 0.55              & 0.45              & 0.60                  & 0.50              & 0.40 \\
\hline
\end{tabular}
\label{Table-Results}
\end{table*}

\begin{figure*}[h]
\centering
\includegraphics[height=10cm,width=17cm]{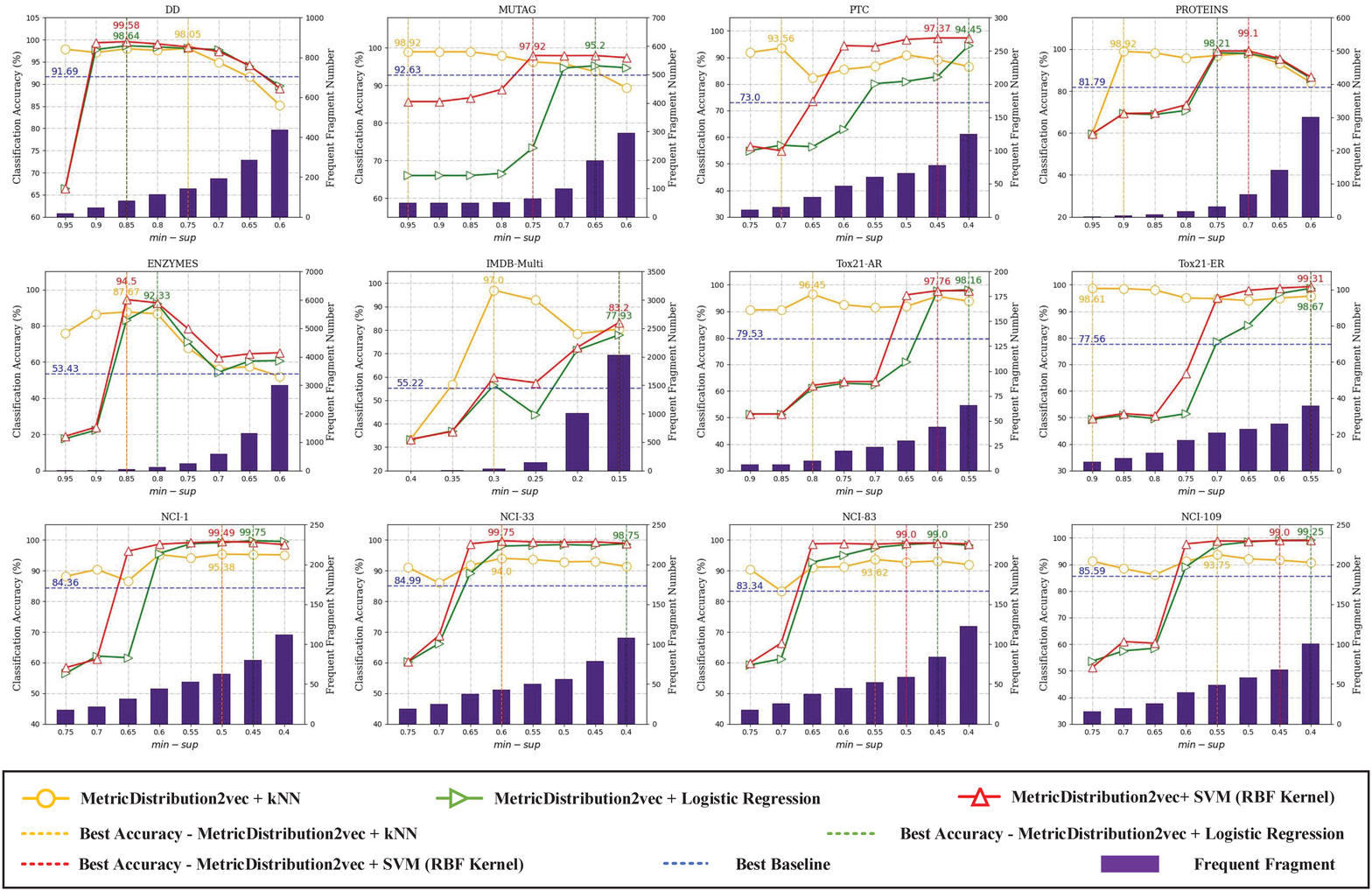}
\caption{The classification accuracy sensitivities of MetricDistribution2vec using kNN, Logistic Regression, and SVM (RBF Kernel) as classifiers over the \emph{min-sup} hyper-parameter are reported with different curves. The blue dotted horizontal line denotes the best result in baselines from Table~\ref{Table-Results}. The different colored vertical lines reflect the best results and the corresponding values of \emph{min-sup} for MetricDistribution2vec using different classifiers. In addition, The number of frequent fragments ($fg$s) under different \emph{min-sup} is shown by the blue histograms.
}
\label{Figure-Sensitivity}
\end{figure*}

In this part, we proceed to the comparison of MetricDistribution2vec with a series of advanced methods on 12 real-world datasets. The overall performances concerning the varying \emph{min-sup} are reported in Figure~\ref{Figure-Sensitivity}. Accordingly, the best classification results of MetricDistribution2vec using kNN, Logistic Regression, or SVM (RBF Kernel) as a lightweight classifier are marked in Table~\ref{Table-Results} with consistent colors of curves in Figure~\ref{Figure-Sensitivity}.

As shown in Table~\ref{Table-Results}, the MetricDistribution2vec using different lightweight classifiers have an apparent strength to contribute to advances in classification precision for all the real-world structural graph datasets. The SVM (RBF Kernel) model as a classifier outperforms other classifiers except on PTC and IMDB-Multi, where the model using kNN behaves much better. On DD, PTC, PROTEINS, Tox21 series, and NCI series datasets, the embedding methods GE-FSG and AWE perform relatively well among the baselines. However, the MetricDistribution2vec + SVM (RBF Kernel) achieves remarkable results on these same datasets, reaching as high as 99.41\% on DD, 97.66\% on PTC, 99.19\% on PROTEINS, 97.63\% on Tox21-AR, 99.36\% on Tox21-ER, 99.38\% on NCI-1, 99.75\% on NCI-33, 99.00\% on NCI-83, and 99.00\% on NCI-109, respectively. On the MUTAG dataset, MetricDistribution2vec + kNN achieves 98.42\% accuracy, which is better than the best baseline resulting from the neural networks model PATCHY-SAN. The ENZYMES and IMDB-Multi datasets are used to verify the multi-classification task. On ENZYMES, the kernel methods, the embedding methods, and the neural network methods all fail to achieve an ideal precision, whose top-ranking accuracy is derived by the Deep Weisfeiler-Lehman equal to around 53.43\%. While MetricDistribution2vec + SVM (RBF Kernel) still promotes the classification accuracy to as high as around 94.50\%. On IMDB-Multi, the MetricDistribution2vec + kNN has a more outstanding precision of 96.67\% than the best result derived from the baselines, which equals 55.22\%.

Figure~\ref{Figure-Sensitivity} is addressed to describe the varying characteristics of accuracy derived from MetricDistribution2vec using three different classifiers with respect to the \emph{min-sup}, which reflects the volume of the selected frequent fragments from the dataset. The models using Logistic Regression and SVM (RBF Kernel) share a monotonically increasing tendency with \emph{min-sup}, which ranges from a large to a small value. This trend implies that a large volume of frequent fragments benefits high classification performance. Conversely, the model using kNN shows little sensitivity to \emph{min-sup} compared with that using Logistic Regression and SVM (RBF Kernel). In particular, the curves on MUTAG, PTC, Tox21 series, and NCI series fluctuate gently in precision. On DD and Enzymes, the curves show a decreasing trend in terms of accuracy after reaching the top points. Notably, the model using kNN prefers a larger \emph{min-sup} value to meet the best classification accuracy on seven datasets, including MUTAG, PTC, PROTEINS, ENZYMES, IMDB-Multi, Tox21-AR, and Tox21-ER.

As an embedding method, MetricDistribution2vec pays more attention to exhibiting the relative relations of each graph with others within the entire set as more as possible, particularly in the way of stacking a series of the quadratic Wasserstein distances associated with the targeted graph into a combined characteristic representation, which is the most notable difference compared to other embedding methods. It is evident that in Table~\ref{Table-Results}, MetricDistribution2vec shows far superior to the compared embedding methods as Graph2Vec, GE-FSG, and AWE on all the 12 datasets, and this firmly confirms the effectiveness of our distinct attempt in graph representation.

\subsection{Visualization for the High-Dimensional Embedded Data}

\begin{figure*}[h]
\centering
\includegraphics[height=9cm,width=17cm]{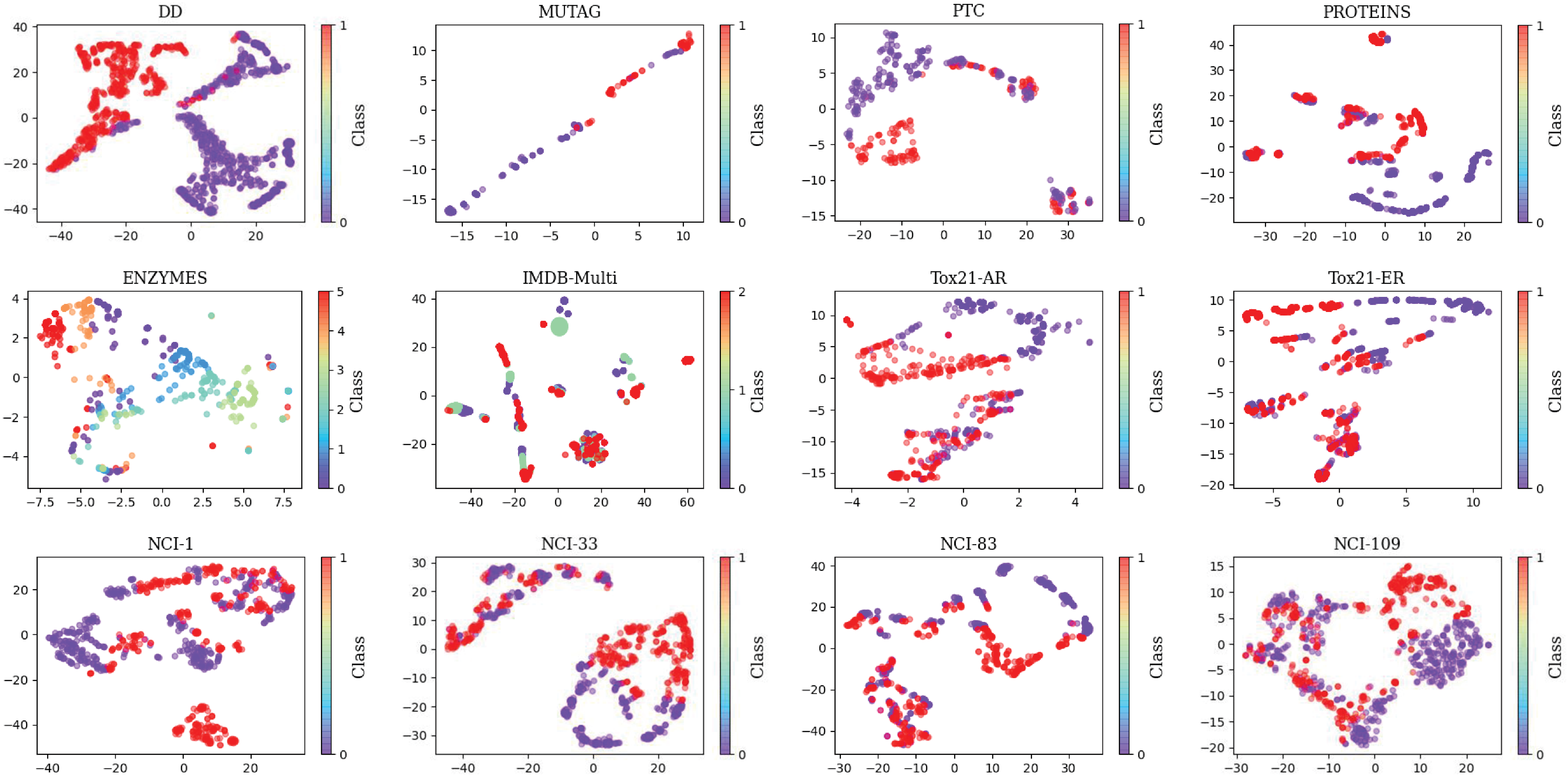}
\caption{The visualization of the high-dimensional embedded data derived from MetricDistribution2vec in a plane by t-SNE for 12 datasets. In each subplot, different colored nodes represent different labeled graphs, and similar embedded graphs are clustered nearby on the plot.
}
\label{Figure-tSNE}
\end{figure*}

As high-dimensional embedded data represents a graph in a vector space, the dimensionality reduction technique like t-Distributed Stochastic Neighbor Embedding (t-SNE)~\cite{tSNE} could be applied to it to visualize the graph in scatter plots. The t-SNE minimizes the Kullback-Leibler divergence~\cite{Kullback-Leibler-divergence} between two probability distributions calculated in the original data space and the embedding space with respect to the points in the embedding. Then it maps different high-dimensional data into other low-dimensional issues according to their similarity in distribution. In Figure~\ref{Figure-tSNE}, embedded graphs from DD, MUTAG, PTC, PROTEINS, Tox21-AR, Tox21-ER, NCI-33, and NCI-83 are separated significantly as there exist fewer overlaps of different colored points in the 2-dimensional plane. While on multi-class datasets ENZYMES and IMDB-Multi, the segmentation boundaries appear a little unclear, and this is consistent with their relatively lower classification accuracies exhibited in Table~\ref{Table-Results}.

\subsection{Experiments In Few-Shot Classification}\label{Experiments-Few-Shot-Classification}

In the above supervised-classification experiments, the proposed MetricDistribution2vec is evaluated by a 10-fold stratified cross-validation way, where $\eta$=90\% objects from the dataset are selected for training, and the left $\zeta$=10\% are used for the test. And the results have verified the high success in data-intensive cases. However, sufficient examples with supervised information are often hard or impossible to acquire in real-world scenarios. Thus, effectively supervised classification based on insufficient training samples, especially the Few-Shot classification, deserves much focus.

\begin{figure*}[h]
\centering
\includegraphics[height=9cm,width=17cm]{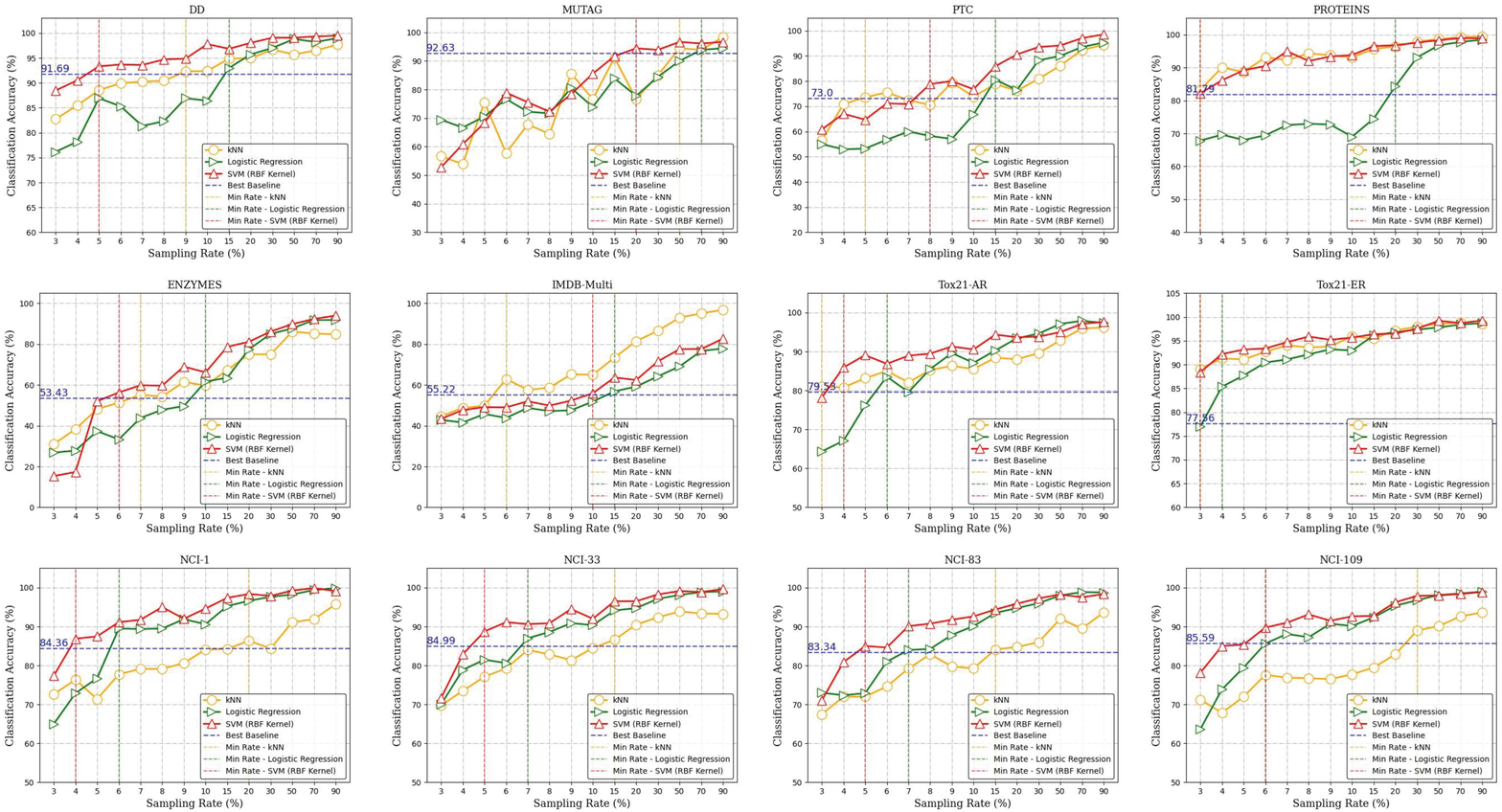}
\caption{The classification accuracy sensitivities of MetricDistribution2vec using kNN, Logistic Regression, and SVM (RBF Kernel) as classifiers over different training data sampling rates are reported in different curves. The blue horizontal dotted line denotes the best result in baselines from Table~\ref{Table-Results}. The different colored vertical dotted lines report the minimum sampling rates that lead to first exceeding the best baseline for different modes and classifiers.
}
\label{Figure-Few-Shot}
\end{figure*}

Few-Shot Learning tends to generalize to new tasks only using prior knowledge learned from a few samples with supervised information. Here we explore the performance of MetricDistribution2vec on Few-Shot classification. In detail, we randomly select a few data points with a small $\eta$ varying from $1\%$ to $90\%$ as the training set for fitting a classifier like kNN, Logistic Regression, or SVM (RBF kernel). We still take $\zeta$=10\% of each database as a test set for validating the learned classifiers. After that, we repeat each sampling and training procedure ten times and report the average predicting accuracy on the test set and the performances of different modes, classifiers, and datasets in Figure~\ref{Figure-Few-Shot}. And in particular, we report the average accuracies calculated by SVM (RBF Kernel) in Table~\ref{Table-Few-Shot} due to its slight advantages in discrimination compared with other classifiers.

In Figure~\ref{Figure-Few-Shot}, we set the blue horizontal dotted line in each subplot to denote the best accuracy of the baselines, which corresponds to the bold blue value in Table~\ref{Table-Results} and Table~\ref{Table-Few-Shot}. The orange, green, and red vertical dotted lines in each subplot reflect the minimum sampling rates that lead to first exceeding the best baselines for kNN, Logistic Regression, and SVM (RBF Kernel), respectively.

We analyze the supervised inference of the MetricDistribution2vec framework from the selection of classifiers. From the global point of view, the SVM (RBF Kernel) exceeds the best baseline with relatively fewer samples, while the Logistic Regression requires more training samples. For the vast majority of the datasets, only $3\%-8\%$ items of the dataset as training samples are enough to induce an accurate SVM (RBF Kernel) to outnumber the best baseline except on MUTAG and IMDB-Binary, where the minimum sampling rates equal at least $20\%$ and $15\%$, respectively. For the Logistic Regression classifier, the minimum sampling rates are still no more than $20\%$ for eleven datasets except on MUTAG.

MetricDistribution2vec is suitable for Few-Shot classification because distribution-based embedding could preserve the inherent metric distribution as much as possible, even if insufficient samples are available for training. Here the volume of the selected training set simply accounts for the approximation of the ground-truth metric distribution but matters little to the overall trend of metric distribution. To measure the discrepancy between two metric distributions on the same graph but derived by different sampling rates, we could still leverage the proposed Wasserstein distance in section~\ref{section-Wasserstein-distance} to calculate the pairwise distance. In detail, for graph $G_{i} \in \mathcal{G}$, suppose its metric distributions under sampling rates $\eta_{1}$ and $\eta_{2}$ are denoted as
\begin{equation}
  \tilde{M}_{G_{i}}^{\eta_{1}} = [\frac{d_{i,1}^{(1)}}{\tilde{\sigma}_{i}^{\eta_{1}}}, \ldots, \frac{d_{i, \tilde{N}_{1}^{\eta_{1}}}^{(1)}}{\tilde{\sigma}_{i}^{\eta_{1}}}, \ldots, \frac{d_{i,1}^{(K)}}{\tilde{\sigma}_{i}^{\eta_{1}}}, \ldots, \frac{d_{i, \tilde{N}_{K}^{\eta_{1}}}^{(K)}}{\tilde{\sigma}_{i}^{\eta_{1}}}]_{1 \times \tilde{N}^{\eta_{1}}},
\end{equation}
and
\begin{equation}
  \tilde{M}_{G_{i}}^{\eta_{2}} = [\frac{d_{i,1}^{(1)}}{\tilde{\sigma}_{i}^{\eta_{2}}}, \ldots, \frac{d_{i, \tilde{N}_{1}^{\eta_{2}}}^{(1)}}{\tilde{\sigma}_{i}^{\eta_{2}}}, \ldots, \frac{d_{i,1}^{(K)}}{\tilde{\sigma}_{i}^{\eta_{2}}}, \ldots, \frac{d_{i, \tilde{N}_{K}^{\eta_{2}}}^{(K)}}{\tilde{\sigma}_{i}^{\eta_{2}}}]_{1 \times \tilde{N}^{\eta_{2}}},
\end{equation}
respectively, the distance between these two metric distributions is given as function~(\ref{Wasserstein-cost-function}) by
\begin{equation}\label{Wasserstein-cost-function-metric-distribution}
  W_{2}(\tilde{M}_{G_{i}}^{\eta_{1}}, \tilde{M}_{G_{i}}^{\eta_{2}})
  = [\inf \limits_{\pi \in \Pi(\tilde{\mu}^{\eta_{1}}, \tilde{\mu}^{\eta_{1}})} \sum_{i = 1}^{ \tilde{N}^{\eta_{1}}} \sum_{j = 1}^{ \tilde{N}^{\eta_{2}}} d^{2}(m^{\eta_{1}}_{i}, m^{\eta_{2}}_{j})] ^{\frac{1}{2}},
\end{equation}
where $m^{\eta_{1}}_{i}$ and $m^{\eta_{2}}_{j}$ denote the $i$-th and $j$-th element in $\tilde{M}_{G_{i}}^{\eta_{1}}$ and $\tilde{M}_{G_{i}}^{\eta_{2}}$.

\begin{figure}[h]
\centering
\includegraphics[height=5.5cm,width=7.5cm]{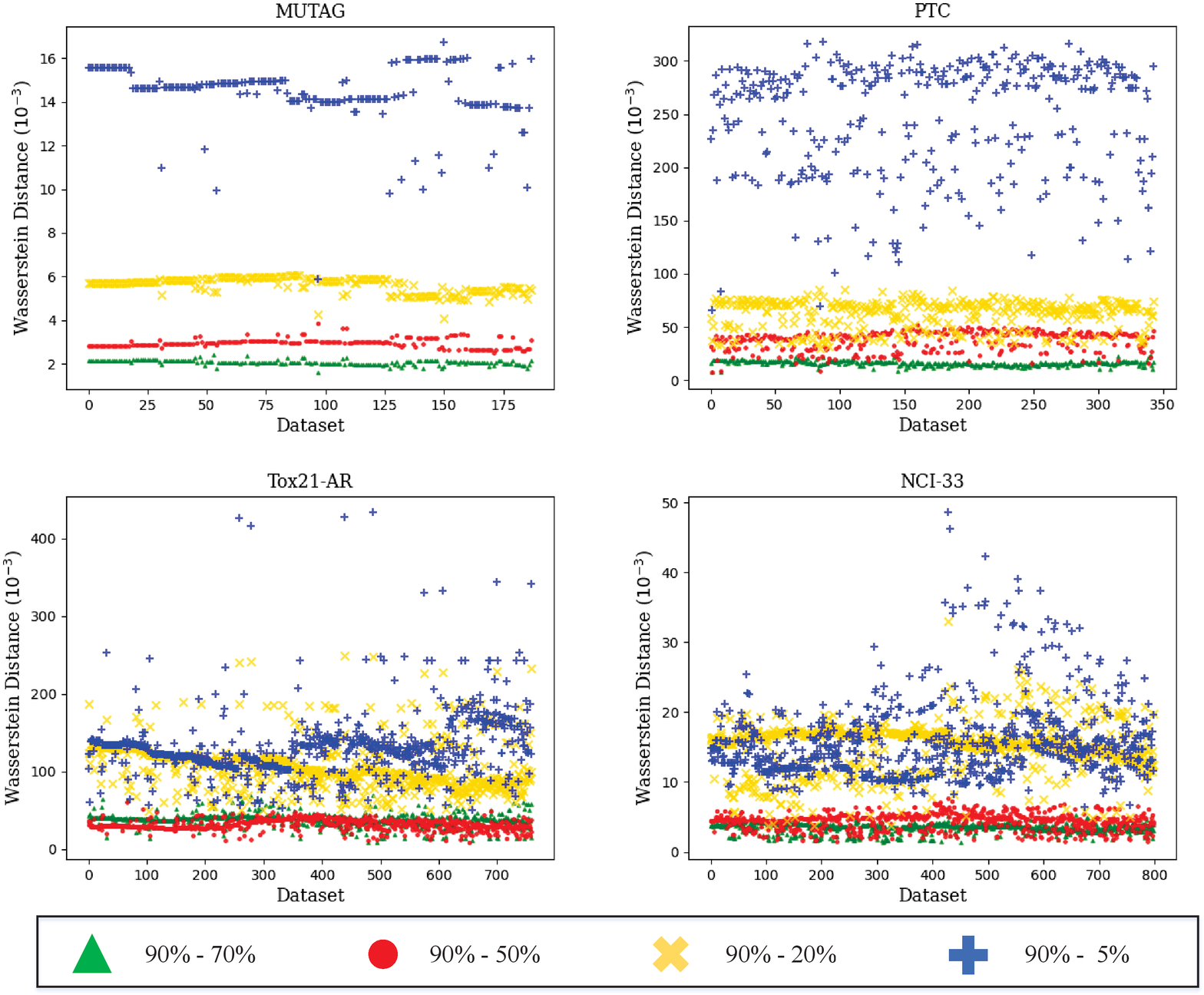}
\caption{This figure shows the similarity between different metric distributions of the same graph under different sampling rates. In each subplot, the horizontal axis denotes the index of each graph, and the vertical axis denotes the distance between different metric distributions. There are four types of symbols on each graph to represent the differences between metric distributions derived by $90\%$ sampling rate and $50\%$, $20\%$, $5\%$ sampling rates, respectively.
}
\label{Figure-MetricDistributionDistance}
\end{figure}

According to Figure~\ref{Figure-MetricDistributionDistance}, we take MUTAG, PTC, Tox21-AR, and NCI-33 as illustration. On every graph data from each dataset, we compute the distance between the metric distribution derived by sampling rate $\eta=70\%$, $50\%$, $20\%$, or $5\%$ and that derived by $\eta=90\%$. Here the $90\%$-deriving metric distribution is taken as the baseline since it reflects the most accurate global metric distribution.

The overall performances of the four subplots are consistent with the results in Table~\ref{Table-Few-Shot} and Figure~\ref{Figure-Few-Shot}. On all the subplots, the red points show more proximity to the green points while the blue points perform much apart from the green points. This indicts that $50\%$-deriving metric distribution preserves most of the baseline metric distribution, and the $5\%$-deriving metric distribution contains relatively little information from the baseline metric distribution. The yellow part, which represents the similarity between $20\%$-deriving metric distribution and the baseline, still shows a small value on all four subplots. This means $20\%$-deriving metric distribution has approximated the $90\%$-deriving metric distribution to a certain extent. Noticing the Tox21-AR and NCI-33 datasets, the blue points exactly overlap the yellow points, meaning the $5\%$-deriving metric distribution is practically the same as the $20\%$-deriving metric distribution. This phenomenon is consistent with the results in Table~\ref{Table-Few-Shot} and Figure~\ref{Figure-Few-Shot}, where the associated classification results are $89.08\%$ and $93.68\%$ for $5\%$-deriving and $20\%$-deriving metric distributions on Tox21-AR, respectively, and $88.75\%$ and $96.50\%$ on NCI-33. While on MUTAG and PTC, the blue points are much far apart from the yellow points due to the classification results equaling $68.34\%$ and $94.44\%$ for $5\%$-deriving and $20\%$-deriving metric distributions on MUTAG, and $64.71\%$ and $90.59\%$ on PTC.

In summary, the MetricDistribution2vec on each dataset, even with the minimum sampling rate and a lightweight classifier, is still superior to the best baseline resulting from 10-fold stratified cross-validation. This result implies that in our proposed framework, a small volume of labeled samples of the database has been enough to carry out the supervised learning tasks and output satisfactory predictions on a series of datasets. Together with the above experimental results and the discussions, we have verified the efficiency and suitability of MetricDistribution2vec on the supervised-classification problems when there is little known about the entire labels of the dataset.

\begin{table*}[htbp]
  \centering
  \caption{Summary of the average classification accuracy (\%) for MetricDistribution2vec using SVM (RBF Kernel) as classifier under different training sample sampling rates. The blue bold values denote the best results in baselines from Table~\ref{Table-Results} and the red bold values report the accuracies as well as the minimum sampling rates that lead to first exceeding the best baseline for different modes on different datasets.}
    \begin{tabular}{|l||c||rrccccccccccc|}
    \hline
    \multicolumn{1}{|l||}{\multirow{2}[1]{*}{\textbf{Dataset}}} & \multicolumn{1}{c||}{\multirow{2}[1]{*}{\textbf{Baseline}}} & \multicolumn{13}{c|}{\textbf{Sampling Rate}} \\
\cline{3-15}          &       & \multicolumn{1}{c}{\textbf{3\%}} & \multicolumn{1}{c}{\textbf{4\%}} & \textbf{5\%} & \textbf{6\%} & \textbf{7\%} & \textbf{8\%} & \textbf{9\%} & \textbf{10\%} & \textbf{15\%} & \textbf{20\%} & \textbf{50\%} & \multicolumn{1}{c}{\textbf{70\%}} & \textbf{90\%} \\
    \hline
    \textbf{DD} & \textcolor[rgb]{ .267,  .447,  .769}{\textbf{91.69}} & 88.46  & 90.51  & \textcolor[rgb]{ 1,  0,  0}{\textbf{93.33 }} & 93.68  & 93.59  & 94.70  & 94.87  & 97.78  & 96.84  & 98.04  & 99.06  & 99.32  & 99.49  \\
    %\midrule
    \textbf{MUTAG} & \textcolor[rgb]{ .267,  .447,  .769}{\textbf{92.63}} & 52.78  & 60.84  & 68.34  & 78.89  & 75.56  & 72.22  & 78.33  & 85.56  & 91.67  & \textcolor[rgb]{ 1,  0,  0}{\textbf{94.44 }} & 96.66  &  96.11  & 96.67  \\
    %\midrule
    \textbf{ENZYMES} & \textcolor[rgb]{ .267,  .447,  .769}{\textbf{53.43}} & 15.34  & 17.33  & 52.00  & \textcolor[rgb]{ 1,  0,  0}{\textbf{56.33 }} & 59.83  & 59.67  & 69.00  & 66.17  & 78.67  & 81.17  & 94.12  &  92.33  & 94.00  \\
    %\midrule
    \textbf{IMDB-Binary} & \textcolor[rgb]{ .267,  .447,  .769}{\textbf{74.45}} & 57.00  & 57.10  & 60.60  & 67.40  & 66.80  & 66.10  & 70.80  & 69.30  & \textcolor[rgb]{ 1,  0,  0}{\textbf{76.30 }} & 79.60  & 88.29  &   90.80    & 92.89  \\
    %\midrule
    \textbf{PTC} & \textcolor[rgb]{ .267,  .447,  .769}{\textbf{73.00}} & 60.82  & 67.06  & 64.71  & 71.18  & 70.88  & \textcolor[rgb]{ 1,  0,  0}{\textbf{78.82 }} & 80.00  & 76.77  & 85.88  & 90.59  & 89.83  & 97.06  & 98.53  \\
    %\midrule
    \textbf{PROTEINS} & \textcolor[rgb]{ .267,  .447,  .769}{\textbf{81.79}} & \textcolor[rgb]{ 1,  0,  0}{\textbf{82.16 }} & 86.13  & 89.19  & 90.45  & 94.96  & 91.98  & 93.33  & 93.78  & 96.49  & 96.76  & 77.53  & 98.92  & 98.83  \\
    %\midrule
    \textbf{Tox21-AR} & \textcolor[rgb]{ .267,  .447,  .769}{\textbf{79.53}} & 78.16  & \textcolor[rgb]{ 1,  0,  0}{\textbf{85.92 }} & 89.08  & 86.84  & 88.95  & 89.47  & 91.32  & 90.53  & 94.34  & 93.68  & 95.00  &  97.10  & 97.63  \\
    %\midrule
    \textbf{Tox21-ER} & \textcolor[rgb]{ .267,  .447,  .769}{\textbf{77.56}} & \textcolor[rgb]{ 1,  0,  0}{\textbf{88.40 }} & 92.19  & 93.15  & 93.37  & 94.71  & 95.88  & 95.19  & 95.67  & 96.37  & 96.53  & 99.20  &  98.72  & 99.25  \\
    %\midrule
    \textbf{NCI-1} & \textcolor[rgb]{ .267,  .447,  .769}{\textbf{84.36}} & 77.38  & \textcolor[rgb]{ 1,  0,  0}{\textbf{86.88 }} & 87.50  & 91.13  & 91.75  & 95.00  & 92.00  & 94.63  & 97.38  & 98.38  & 99.25  & 99.88  & 99.13  \\
    %\midrule
    \textbf{NCI-33} & \textcolor[rgb]{ .267,  .447,  .769}{\textbf{84.99}} & 71.63  & 82.88  & \textcolor[rgb]{ 1,  0,  0}{\textbf{88.75 }} & 91.13  & 90.63  & 90.88  & 94.50  & 92.00  & 96.50  & 96.50  & 99.13  & 98.88 & 99.63  \\
    %\midrule
    \textbf{NCI-83} & \textcolor[rgb]{ .267,  .447,  .769}{\textbf{83.34}} & 71.00  & 80.88  & \textcolor[rgb]{ 1,  0,  0}{\textbf{85.00 }} & 84.63  & 90.13  & 90.75  & 91.75  & 92.63  & 94.38  & 95.88  & 98.25  & 97.50  & 98.38  \\
    %\midrule
    \textbf{NCI-109} & \textcolor[rgb]{ .267,  .447,  .769}{\textbf{85.59}} & 78.13  & 85.00  & 85.38  & \textcolor[rgb]{ 1,  0,  0}{\textbf{89.75 }} & 91.00  & 93.13  & 91.50  & 92.63  & 92.75  & 96.25  & 98.00  & 98.38 & 98.88  \\
    \hline
    \end{tabular}%
  \label{Table-Few-Shot}%
\end{table*}%

\section{Conclusion and Future Outlook}\label{Conclusion-and-Future-Outlook}

The metric distribution at each targeted graph has provided a broad-scale perspective to better understand the inherent discrepancy apart from others within the entire dataset. This paper embeds metric distribution into the vectorial representation for structural graph data and proposes its straightforward applications in both the supervised classification and Few-Shot classification tasks. The worthy mention of this framework differing from pioneering embedding works lies in the independent attempt for graph representation. It aims to characterize every graph with a series of pairwise similarities within the global scale rather than a description of every graph's inner properties. To quantitatively measure the similarity between structural graphs, this paper presents a novel distance metric using the optimal transportation theory. In detail, the optimal transportation is built to bridge the pairwise graphs characterized basically by their generation distributions, exhibited as decompositions of a series of substructures. In particular, we focus on a unique form of substructures referred to as frequent fragments, according to a preset sampling threshold hyper-parameter  \emph{min-sup}, to reflect the commonness of crucial components within the entire dataset.

The main advantages of our proposed method are the high precision of supervised classification in both ten-fold cross-validation and Few-Shot scenarios, and the suitability for lightweight classifiers. In the classification experiments with ten-fold cross-validation data partition, our method could far exceed all the published baselines on all datasets, with the highest classification accuracy reaching $99\%$. In the Few-Shot experiments with insufficient training samples, for the vast majority of the datasets, $3\% - 8\%$ of the data as the training samples are enough to learn a well discriminative classifier to get the same effort of the corresponding best baseline. In addition, the results show the classification insensitive to the selection of classifiers. This fact verifies the feasibility of our method for particular scenes that demand lightweight algorithms.

The proposed representation strategy of this paper has provided an easy but practicable way to effectively distinguish objects with blurred decision boundaries. Potential applications of this strategy could be expected in a broader range of pattern recognition tasks related to metric learning, for example, image recognition, target identification under disturbance, etc. As the pairwise distance computing algorithm used in this paper has not been designed for efficiency, we will also consider much room for improvement in the calculation in future work. In addition, the optimal transportation measure for the similarity between the evolving data with respect to varying times would be another attractive topic for further study.

\section*{Acknowledgements}
This work is supported by the Research and Development Program of China (Grant No. 2018AAA0101100), the National Natural Science Foundation of China (Grant Nos. 62276013, 62141605, 62050132), the Beijing Natural Science Foundation (Grant Nos. 1192012, Z180005).

\bibliographystyle{unsrt}
\bibliography{StructuredDataUsingOT_bibfile.bib}

\end{document}